\documentclass[letterpaper, 10 pt, conference]{ieeeconf}

\IEEEoverridecommandlockouts                              

\usepackage{epsfig} 
\usepackage{mathrsfs}

\usepackage{times} 
\usepackage{amsmath} 
\usepackage{amssymb}  
\usepackage{bm}

\usepackage{cite}
\usepackage{color}
\usepackage{xcolor}
\usepackage{xspace}
\usepackage{array}

\usepackage{algorithm}
\usepackage{algorithmic}

\usepackage{graphicx}
\usepackage{subcaption}
\usepackage{mwe}
\usepackage{comment}
\usepackage{url}

\usepackage[labelsep=endash]{caption}

\usepackage[colorlinks,bookmarksopen,bookmarksnumbered,citecolor=black,urlcolor=blue,linkcolor=purple]{hyperref}

\newtheorem{theorem}{Theorem}[section]
\newtheorem{assumption}{Assumption}[section]
\newtheorem{lemma}{Lemma}[section]

\usepackage{setspace}
\title{\LARGE
Bayesian Online Learning for Human-assisted Target Localization
}

\author{Min-Won Seo$^{1}$ and  Solmaz S. Kia$^{1}$, \emph{Senior Member, IEEE}
\thanks{$^{1}$Min-Won Seo, and Solmaz S. Kia are with the Department of Mechanical and Aerospace Engineering,
        University of California, Irvine, CA 92697, USA,
        {\tt\small \{minwons,kia\}@uci.edu}}%
}

\newcommand{\real}{{\mathbb{R}}}

\newtheorem{remark}{Remark}[section]

\begin{document}\fontsize{10}{11.13}\rm

\maketitle
\thispagestyle{empty}
\begin{abstract}
    We consider a human-assisted autonomy sensor fusion for dynamic target localization in a Bayesian framework. Autonomous sensor-based tracking systems can suffer from observability and target detection failure. Humans possess valuable qualitative information derived from their past knowledge and rapid situational awareness that can give them an advantage over machine perception in many scenarios. To compensate for the shortcomings of an autonomous tracking system, we propose to collect spatial sensing information from human operators who visually monitor the target and can provide target localization information in the form of free sketches encircling the area where the target is located. However, human inputs cannot be taken deterministically and trusted absolutely due to their inherent subjectivity and variability. Our focus in this paper is to construct an adaptive probabilistic model for human-provided inputs where the adaptation terms capture the level of reliability of the human inputs. The next contribution of this paper is a novel joint Bayesian learning method to fuse human and autonomous sensor inputs in a manner that the dynamic changes in human detection reliability are also captured and accounted for. Unlike deep learning frameworks, a unique aspect of this Bayesian modeling framework is its analytical closed-form update equations. This feature provides computational efficiency and allows for online learning from limited data sets. Simulations demonstrate our results, underscoring the value of human-machine collaboration in autonomous systems.
\end{abstract}

\medskip
\section{Introduction}
\label{sec::intro}
\begin{figure}[!t]
    \centering
    \includegraphics[width=0.44\textwidth]{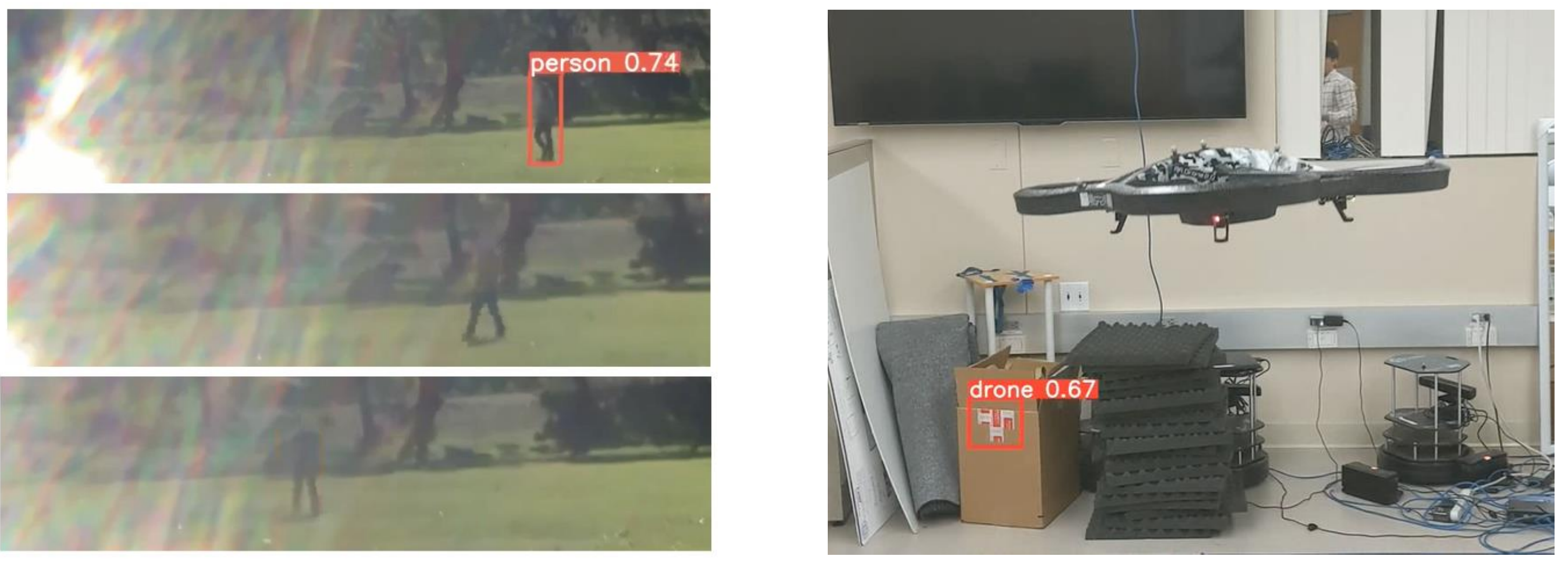}
    \caption{\small{Examples of the machine learning-based object detection algorithm (YOLO~\cite{farhadi2018yolov3}) failure to detect (a person on the left figures) or miss-detect (a drone on the right figure). (Left) The initial correct detection is compromised due to intense sun glare. (Right) Detection falters due to a complex background and insufficient training data. In both these examples, a human can readily identify the targets, providing valuable assistance to tracking systems.}}
    \label{fig:macine_fail}
\end{figure}

Dynamic target localization using sensor measurements collected by autonomous agents has been of interest for a long time. Multi-agent target localization where multiple agents from different perspectives obtain measurements from the target and fuse them together has demonstrated an improvement in accuracy~\cite{stachura2011cooperative,nagaty2015probabilistic,robin2016multi}. Despite advances in sensor data fusion algorithms and autonomous perception abilities, relying merely on the hard sensor measurements collected by only the autonomous sensors does not always result in an acceptable level of localization accuracy in complex environments. This is because autonomous sensors can suffer from observability and target recognition failures, see Fig.~\ref{fig:macine_fail}. Recently, including human observers that can provide spatial information to localize dynamic targets in challenging environments has been proposed to compensate for the shortcomings of the autonomous sensors~\cite{kaupp2007shared,kennedy2007spatial,burks2018closed,muesing2021fully}. This has resulted in effective tracking and localization of the target in time-critical tasks. 

Humans possess valuable qualitative information derived from their past knowledge and rapid situational awareness that can give them an advantage over machine perception in many scenarios. However, it is essential to admit that sensory inputs provided by humans are inherently \emph{subjective and soft} in nature~\cite{hall2010human}. Humans are prone to making mistakes, and their abilities can vary depending on the situation, prior experience, and even the current state of mind. Consequently, excessive reliance on human input can lead to poor performance and should be avoided~\cite{sheridan2002humans}. It is critical to have a proper \emph{mathematical model} for quantifying human inputs to ensure the success and effectiveness of human-assisted autonomy for localization tasks. The model should consider uncertainty factors such as human reliability (errors). Additionally, the model parameters need to be updated online to capture changes in the uncertainty of human reliability over time. By considering these aspects, we can build estimation pipelines that enable efficient integration of human input, thus facilitating improved localization outcomes in human-assisted autonomy systems. 

In recent years, human-assisted autonomy sensor fusion has played a significant role in various sensing and planning tasks. For example, in~\cite{kunze2014using}, qualitative spatial constraints for manipulators are acquired from human spoken commands. However, variations in human pronunciation and ambient noise levels can impact the system's effectiveness. On the other hand,~\cite{robinson2015human} proposes a fusion of human electroencephalography (EEG) and autonomic sensor data, thus requiring additional hardware and accounting for EEG sensitivity to external interference. \cite{cai2018human} primarily focuses on human-assisted site inspection tasks aimed at addressing real wireless communication, rather than emphasizing human information expression. In~\cite{pavliv2021tracking}, the authors present a method to detect and track targets from a human perspective using a headset, which might be affected by the human's field of view. These studies highlight the valuable contributions of human-provided information in collaborative tasks that cannot be obtained solely from autonomous systems. In this paper, our focus is on the use of human sketch-based spatial localization in dynamic target tracking.

In dynamic target localization, sensor information plays a crucial role in continuously tracking the target. Additionally, information should be presented in a clear way (reduced language barriers) and concisely (data compression). Leveraging drawing-type observations as a semantic language interface presents potential advantages. First, it can be readily applied to widely used mobile systems, such as smartphones, tablets, and emerging technologies like Augmented Reality (AR) and Virtual Reality (VR). Secondly, drawing-based information is highly intuitive and practical for both robots and humans. However, the observation model must proficiently capture human reliability \emph{without} imposing a computational burden while delivering accurate information. Research on human drawing information fusion is conducted within robotics applications~\cite{shah2013qualitative,ahmed2015fully,burks2023harps}. The authors of~\cite{shah2013qualitative} illustrate humans conveying map data via hand-sketched drawings for robot navigation, however, neglect inherent human-provided information uncertainty. In contrast,~\cite{ahmed2015fully} presents a human sketch-based method for static target search with uncertainty, but it doesn't consider the computational costs of parameter updates. The authors of~\cite{burks2023harps} place their primary emphasis on online planning using human drawing observations, rather than focusing on updating human parameters.

\smallskip
\emph{Statement of contribution}: This paper introduces a novel human drawing model incorporating uncertainty for dynamic target localization, emphasizing the importance of computational efficiency in updating the distribution of human parameters (detection reliability). Specifically, the distribution of human parameters is updated in the \emph{closed-form} through Bayesian learning, enabling real-time tracking tasks. First, we present a novel probabilistic observation model for human drawings that considers not only the reliability of human detection but also incorporates inherent uncertainty. Moreover, its conjugate closed-form structure enhances computational efficiency, a crucial factor in dynamic target localization. In addition, our next contribution is to propose a joint Bayesian learning approach that concurrently localizes the target and updates human parameters. This strategy enables online adjustments of human parameters to precisely depict the evolving human detection reliability over time. Finally, we demonstrated improved target localization results and the updated distribution of human detection reliability in a simulation study.
  
The remainder of this paper is organized as follows: Section~\ref{sec::ProbDef} presents the problem definition. Section~\ref{sec::TLFramework} describes the particle-based target localization framework. In Section~\ref{sec::human_obs}, we present our novel human-drawing observation likelihood model, which incorporates human detection reliability. Furthermore, in Section~\ref{sec::BayesLearning}, we discuss joint Bayesian learning for target localization and human parameters. Section~\ref{sec::simulation} reports the simulation results. Finally, Section~\ref{sec::Con} gives the conclusion.

\begin{figure}[!t]
    \centering
    \includegraphics[width=0.40\textwidth]{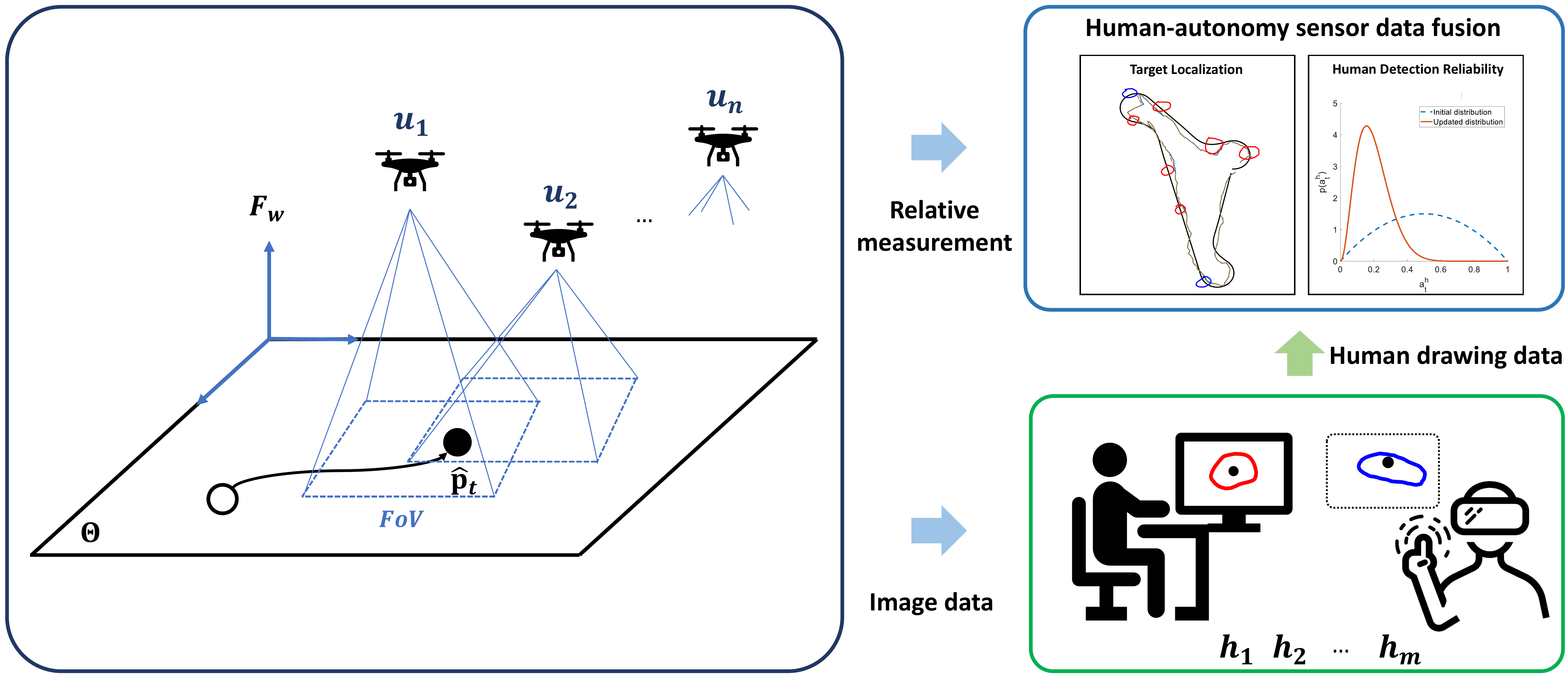}
    \caption{{\small A representative scenario of the problem of interest in this paper: A target, depicted by $\bullet$, moves in a 2D space. UAVs equipped with a stereo vision camera transmit relative measurements to a centralized fusion center, and image data to human operators. The human operators provide inside drawing observations on the image data via a touch screen monitoring system (e.g., tablet). The human and autonomous sensor data fusion is performed to localize the target in the centralized system.}}
    \label{fig:HMM_Loc}
\end{figure}

\medskip
\section{Problem Definition}
\label{sec::ProbDef}
This paper considers a dynamic target localization problem in which a team of autonomous sensors $u \in \mathcal{U}= [u_1,\cdots,u_n]^\top$ with help from human operators $h \in \mathcal{H}= [h_1,\cdots,h_m]^\top$ estimate the location of a mobile target, see Fig.~\ref{fig:HMM_Loc}. The target moves on a 2D terrain $\Theta \subset \real^2$. The sensors, e.g., a stereo camera with a limited field of view, provide image data of the target and relative range information between the target and themselves; measurements can also be other types of observations, e.g., bearing measurement. We assume that real-time, possibly low-resolution, footage of the target is streamed for human observers, e.g., via a UAV flying over the scene~\cite{guo2020design}. Human operators offer spatial sensing about the target's location based on their visual perception of the image data by enclosing the location of the target on the image plane. The observations are transmitted to a centralized system for integrating the data and estimating the target position. The observations are based on the following assumption:
\begin{assumption}\label{assum::cond}
    The observations at time $t$ by sensors and human observers are under independent and identically distributed (i.i.d) conditions, and there is no conditional dependency between them due to no information exchange among team members.
\end{assumption}

\smallskip
Let the position of the target at time $t$ be denoted by $\mathbf{p}_t=[x_t~y_t]^\top \in \Theta$, where $x_t$ and $y_t$ represent its 2D coordinates at time index $t$ in a 3D world frame $\mathcal{F}_w$. When the target position vector is expressed in 3D coordinates, it is represented as $\mathbf{p}_t^+ = [\mathbf{p}_t^\top~0]^\top$. The state of the sensing platform $u$ is the camera pose $\mathbf{s}_{u,t}= [\mathbf{c}_{u,t}^\top~ \psi_{u,t}]^\top$, where $\mathbf{c}_{u,t} \in \real^3$ is the position of the sensor, referred to $\mathcal{F}_w$, and $\psi_{u,t} \in [0, 2\pi]$ is the yaw angle of the sensing platform. We assumed image data is obtained in the face-down view~\cite{sun2016camera}, however, it can be extended to any view of the image data. Our objective is to improve the accuracy of estimating the target's position, denoted as $\hat{\mathbf{p}}_t$, by modeling and online learning from human operators' observations.

\section{Target Localization Framework}
The likelihood model for range observation from autonomous sensors is a nonlinear function of the target state, and human observation is a non-Gaussian model, resulting in complex posterior distributions. To solve our problem, thus, we propose to use a finite-state Hidden Markov Model (HMM) based framework. This non-parametric Bayesian framework offers the advantage of expressing probabilities explicitly using a discrete sample space, enabling recursive Bayesian inference in \emph{closed} form. Additionally, it seamlessly incorporates geometric information, such as barriers or boundaries, into localization~\cite{rudic2020geometry}. Furthermore, integration with the control and planning pipeline is readily achieved, as finite state-based techniques commonly establish action (control) policies~\cite{rosolia2023model}.

\begin{figure}[!t]
    \centering
    \includegraphics[width=0.43\textwidth]{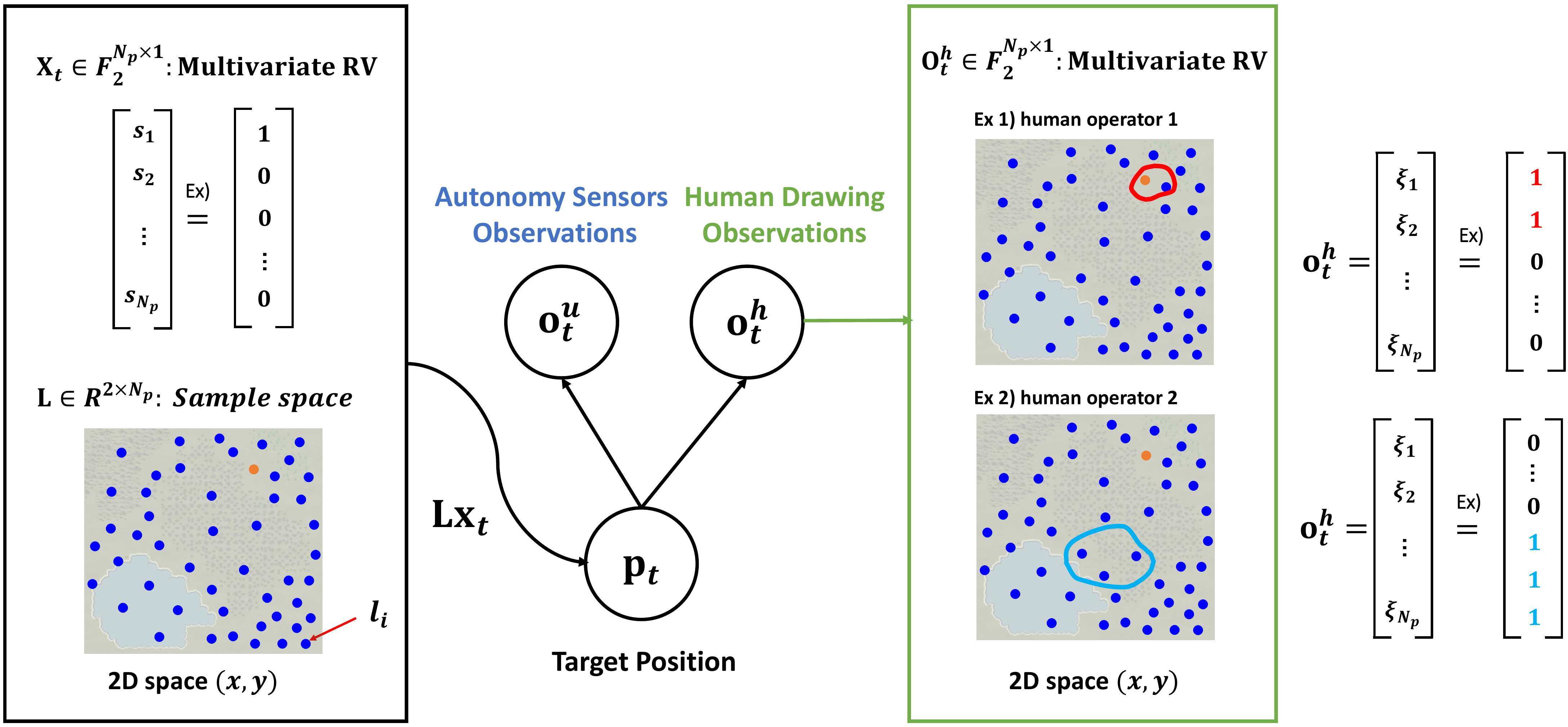}
    \caption{{\small $\mathbf{X}_t \in \mathbb{F}_2^{N_p \times 1}$ represents a multivariate random variable at every time step. Each element $s_i$ of $\mathbf{X}_t$ is mapped into the particle $l_i$ in 2D sample space $\mathbf{L} \in \mathbb{R}^{2 \times N_p}$. There is human-drawing observation $\mathbf{O}_t^h \in \mathbb{F}_2^{N_p \times 1}$, indicating the target inside. In this example, two human operators draw the region, where, unlike $\mathbf{X}_t$ that only has a single $s_i = 1$, $\mathbf{O}^h_t$ can have multiple $\zeta_j = 1$.}}
    \label{fig:HMM}
\end{figure}

\smallskip
\label{sec::TLFramework}
\emph{Hidden Markov Model Framework}:
We discretize the target workspace via $N_p$ particles which generate a 2D sample space $\mathbf{L} = [\mathbf{l}_1, \dots, \mathbf{l}_{N_p}] \in \mathbb{R}^{2 \times N_p}$ where $\mathbf{l}_i =[x~y]^{\top} \in \Theta$ is the position of $i$-th particle and $N_p$ is the total number of particles, see Fig.~\ref{fig:HMM}. The HMM has two stochastic processes, hidden states, $\mathbf{x}_t \in \mathbf{X}_t$, and observations, $\mathbf{o}^u_t \in \mathbf{O}^u_t$, $\mathbf{o}^h_t \in \mathbf{O}^h_t$ at each time step $t$. The hidden state encapsulates the belief regarding the target's position within a discretized sample space. Observations are obtained from autonomous sensors and human operators. The hidden state $\mathbf{x}_t$ takes values from $\mathbf{X}_t = [s_1,s_2,\dots,s_{N_p}]^{\top} \in \mathbb{F}_2^{N_p \times 1}$ where $s_i$ takes the binary number $1$ or $0$. Each element $s_i$ is mapped one-to-one into the particle $\mathbf{l}_i$ in the particle-based sample space. Since we are localizing only one target, $\mathbf{X}_t$ can have \emph{only one} $s_i=1$, and \emph{all the other elements are set to} $s_j\!=\!0, \; j \neq i$. Therefore, multiplication $\mathbf{L} \mathbf{x}_t$ corresponds to the position of the target $\mathbf{p}_t$ ($\mathbf{p}_t = \mathbf{L} \mathbf{x}_t$). While autonomous sensors' range observation $\mathbf{o}^u_t$ can take real values $\mathbf{O}^u_t \in \real_{\geq 0}$, human-drawing observation $\mathbf{o}^h_t$ takes values from $\mathbf{O}^h_t = [\zeta_1,\zeta_2,\dots,\zeta_{N_p}]^\top$,  $\mathbf{O}^h_t \in \mathbb{F}_2^{N_p \times 1}$ where $\zeta_i$ takes the binary number $1$ or $0$. Unlike $\mathbf{X}_t$, which only has a single $s_i = 1$ at each time step, $\mathbf{O}^h_t$ can have \emph{multiple} $\zeta_j = 1$, as depicted in Fig.~\ref{fig:HMM}. Therefore, it is important to carefully quantify the information between $\mathbf{X}_t$ and $\mathbf{O}^h_t$. It will be discussed in Section~\ref{sec::human_obs}.

\smallskip
\emph{Forward Algorithm for Recursive Bayesian Filtering}:  
Our goal is to find the posterior distribution of the target’s state, given all observations $\{\mathbf{o}_{1:t}^u, \mathbf{o}_{1:t}^h\} \subset \mathbf{o}_{1:t}$, as given by
\begin{align}
\label{eq::Traj_Est}
    & p(\mathbf{p}_t|\mathbf{o}_{1:t}) \!=\!\eta \,p(\mathbf{o}_t|\mathbf{p}_t) \!\! \int_{\Theta} \!\! \!p(\mathbf{p}_t|\mathbf{p}_{t-1}) p(\mathbf{p}_{t-1}|\mathbf{o}_{1:t-1}) \text{d}\mathbf{p}_{t-1},
\end{align}
where $\eta$ is a normalizing constant that ensures the posterior adds up to $1$. A prior distribution $p(\mathbf{p}_0)$ is constructed from any information available a prior. Here, $p(\mathbf{p}_{t-1}|\mathbf{o}_{1:t-1})$ is a posterior distribution at the previous step $t-1$, $p(\mathbf{p}_t|\mathbf{p}_{t-1})$ is a Markovian state transition model for the target. The integral term is from Chapman-Kolmogorov equation~\cite{papoulis2002probability}. $p(\mathbf{o}_t|\mathbf{p}_t)$ is an observation model (referred to as likelihood function in the probabilistic form).

Among well-known HMM-based algorithms~\cite{svensen2007pattern}, we use the forward algorithm to estimate $p(\mathbf{p}_t | \mathbf{o}_{1:t})$. The forward algorithm shares similarities with the Kalman filter, as it involves prediction and correction steps. However, the fundamental distinction lies in the nature of the sample space, which is discrete (finite) rather than continuous (infinite). The integral term and $\eta$ in~\eqref{eq::Traj_Est} can be calculated through summation. As a result, the posterior distribution of target positions is approximated with a set of particles with corresponding weights $\{\mathbf{p}^i_t, q^i_{t|t}\}_{i=1}^{N_p}$ as given by
\begin{align}
\label{eq::Post_Target_Approx}
    p(\mathbf{p}_t|\mathbf{o}_{1:t}) = \sum\nolimits_{i=1}^{N_p} q^i_{t|t} \cdot \delta(\mathbf{p}_t - \mathbf{p}^i_t),
\end{align}
where $\delta(\cdot)$ is the Dirac delta function, and $q^i_{t|t} \in \real_{\geq0}$ is the weight for the $i^{th}$ particle $\mathbf{p}^i_t$ and satisfies $\sum_{i=1}^{N_p} q^i_{t|t} = 1$. The update of $q^i_{t|t}$ involves two steps, denoted as
\begin{subequations} \label{eq::Post_update}
\begin{align}
    q^i_{t|t-1} &= \sum\nolimits_{j=1}^{N_p} p(\mathbf{p}^i_t|\mathbf{p}_{t-1}^j) q^j_{t-1|t-1},  \quad~ \mbox{\text{(prediction)}}, \label{eq::Post_update1}\\
    q^i_{t|t} &= \eta~p(\mathbf{o}_t|\mathbf{p}_t^i) q^i_{t|t-1}, \qquad\qquad~~~~~\mbox{\text{(correction)}}\label{eq::Post_update2},
\end{align}
\end{subequations}
where $\eta$ ensures $\sum\nolimits_{i=1}^{N_p}p(\mathbf{o}_t|\mathbf{p}_t^i) q^i_{t|t-1} = 1$ and $q^j_{t-1|t-1} = p(\mathbf{p}_{t-1}^j|\mathbf{o}_{1:t-1})$. Subsequently, the minimum mean square error (MMSE) estimate of the position of the target is computed by $\hat{\mathbf{p}}_t \!=\! \int_{\Theta}\! \mathbf{p} \cdot p(\mathbf{p}_t|\mathbf{o}_{1:t}) \text{d}\mathbf{p} \approx\! \sum\nolimits_{i=1}^{N_p} \!\mathbf{p}^i_t \cdot q^i_{t|t}$.

\medskip
\section{The Human Drawing Likelihood Model}\label{sec::human_obs}
In the context of human-drawn spatial observations, we consider \emph{inside-drawing} observations, which indicate the presence of the target inside an area (see Fig.~\ref{fig:HMM}).\footnotemark{}\footnotetext{Assuming the human drawing is a simplex closed-area shape, the drawing on the visual interface plane (e.g., tablet) is approximated as a set of finite distinct points. Then, these points are transformed into the 2D plane $\Theta$ using a rotation matrix $\mathbf{R}_{w}^{c}(\psi_{u,t})$ and the camera intrinsic parameter matrix $\mathbf{K}$~\cite{hartley2003multiple}. This transformation enables the calculation of the human drawing likelihood in the 2D space $\Theta$.} The ability of human operators to detect the target in images varies among individuals and can be influenced by factors such as their mental state, workload, and environmental conditions. Our proposed human drawing observation model accounts for these variations in detection reliability, considering changes in human operators' reliability over time.

\begin{figure}[!t]
    \centering
    \includegraphics[width=0.41\textwidth]{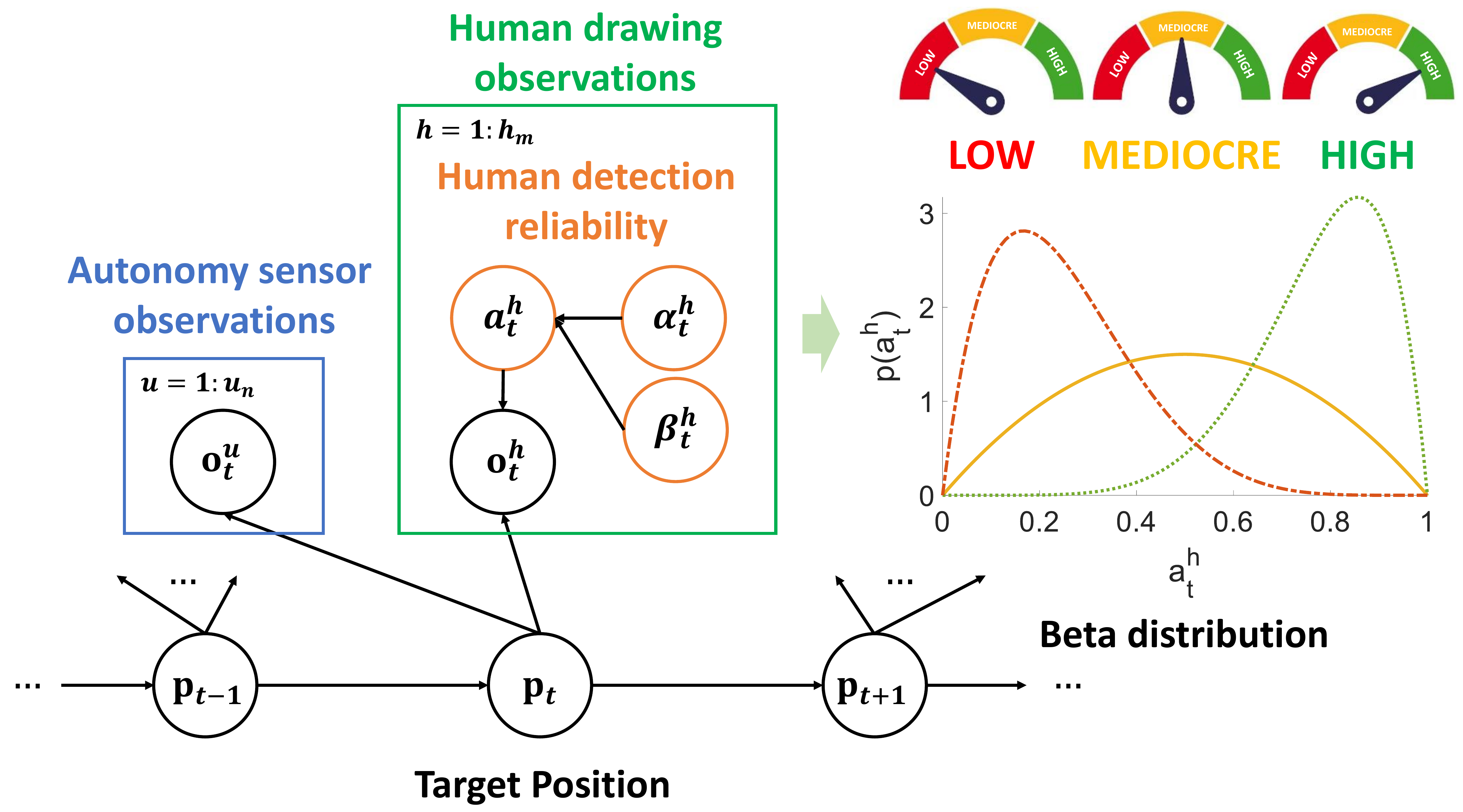}
    \caption{{\small Probabilistic graph model for human-assisted autonomy sensor fusion. $\mathbf{p}_t$ is the position of a target at a given time $t$. Autonomy sensors observations $\mathbf{o}_t^u \in \mathbf{O}_t^u$ are obtained from mobile agents $u \in \mathcal{U}$ at a given time $t$. Human-drawing observations $\mathbf{o}_t^h \in \mathbf{O}_t^h$ are provided by human operators $h \in \mathcal{H}$ at a given time $t$. At time index $t$, the human $h$'s detection reliability for drawing observations is represented as $a^h_t \sim \mathsf{Beta}(\alpha_t^h,\beta_t^h)$. By using different values for $\alpha_t^h$ and $\beta_t^h$, we can capture various levels of reliability.}}
    \label{fig:PGM}
\end{figure}

\medskip
\emph{Prior Model for Human Detection Reliability}:
To design the human drawing model, we represent human detection reliability for human-drawing observations as conditional probability. Let's consider the \emph{ground truth} state of the target, $\mathbf{x}_t^g \in \mathbf{X}_t$, and the human drawing observation $\mathbf{o}_t^h \in \mathbf{O}_t^h$ where only one element $\zeta_i$ is set to $1$. Given the \emph{ground truth} target state $\mathbf{x}_t^g$, the probability of the inside drawing observation can be expressed by
\begin{align}
\label{eq::Human_Inside} 
    p(\mathbf{O}_t^h\!=\!\mathbf{o}_t^h|\mathbf{X}_t \! =\! \mathbf{x}_t^g)  \!=\! \left\{  
        \begin{array}{ccc} 
         \!  \! a^h_t,  & \mathbf{o}_t^h \!=\! \mathbf{x}_t^g~~~(\mbox{true})\\ 
           \! \!1- a^h_t,  &~ \mathbf{o}_t^h \neq \mathbf{x}_t^g~~~(\mbox{false}) 
        \end{array} \right.\!\!, 
\end{align}
where $a^h_t \in [0,1] \subseteq \real$ indicates human detection reliability at time $t$. A high value of $a^h_t$ indicates strong trust in human input (high reliability), and conversely, a low value indicates less trust (low reliability). However, human abilities are different from person to person and can also change during the course of an operation. Therefore, to quantify the uncertainty of the human detection reliability, we model it as a random variable with Beta distribution,  $a^h_t \sim \mathsf{Beta}(\alpha_t^h,\beta_t^h)$, where $\mathsf{Beta}(.,.)$ denotes the Beta distribution and $\alpha_t^h ,\beta_t^h\in \real_{>0}$  are the parameters of the distribution at time $t$. With a Beta distribution, the probability density~function of $a^h_t$ is
\begin{align}\label{eq::Human_param_dist_density}
p(a^h_t) = 1/\mathsf{B}(\alpha_t,\beta_t)\cdot(a^h_t)^{\alpha_t -1} \cdot (1-a^h_t)^{\beta_t -1},
\end{align}
where $\mathsf{B}(\alpha_t,\beta_t)=\Gamma(\alpha_t)\Gamma(\beta_t)/\Gamma(\alpha_t+\beta_t)$ is the Beta function, a normalization constant to ensure that the probability is $1$, and $\Gamma(\cdot)$ is the Gamma function~\cite{wackerly2014mathematical}. The Beta distribution, defined within the interval $[0,1] \subseteq \real$, is well-suited for statistical modeling of human reliability~\cite{guo2021modeling}. For different values of $\alpha_t^h$ and $\beta_t^h$, we can capture different levels of reliability as shown in Fig.~\ref{fig:PGM} on the right (low, mediocre, high). For simple notation, we will omit the subscript $h$ from $\alpha_t^h$ and $\beta_t^h$, except when it is necessary. We will learn these parameters as part of our Bayesian learning method.

\medskip
\emph{Likelihood Model for Human Drawing Observation}: 
Based on the human detection reliability $a^h_t$, we propose a likelihood function for human-drawing observations. Initially, we will focus on $\mathbf{o}_t^h$, where \emph{`only one'} element $\zeta_i$ is set to $1$, while all other elements $\zeta_j$ are set to $0$ ($j \neq i$). Given $\mathbf{X}_t = \mathbf{x}_t$, the likelihood function for inside drawing $\mathbf{o}_t^h$ of human operator $h$ can be expressed using the Kronecker delta function $\delta (\cdot, \cdot)$ by
\begin{align}
\label{eq::Human_Inside_likelihood_one}
     p(\mathbf{o}_t^h|\mathbf{x}_t, a^h_t; \mathbf{L}) \!=\!(a^h_t)^{\delta (\mathbf{x}_t, \mathbf{o}_t^h)} 
     \!\times\!(1\!-\!a^h_t)^{\bigl(1\! - \delta (\mathbf{x}_t, \mathbf{o}_t^h)\bigl)},
\end{align}
where $\delta (\mathbf{x}_t, \mathbf{o}_t^h) = 1$ if $\mathbf{x}_t =\mathbf{o}_t^h$, and $\delta (\mathbf{x}_t, \mathbf{o}_t^h) = 0$ otherwise. Next, we extend our consideration to include \emph{`multiple'} $\zeta_i=1$ in $\mathbf{o}_t^h$ (e.g., $\mathbf{o}_t^h = [1, 1, 0,\dots, 0]^\top$). Let's define a set $\mathcal{O}_t =\bigl\{ [\xi_1, \xi_2, \dots, \xi_{N_p}]^\top \in \mathbb{F}_2^{N_p \times 1} | \sum_{\forall i} \xi_i = 1, i \in \{1,\dots,N_p\}\bigl\}$. When the human drawing observation $\mathbf{o}_t^h$ contains $M_t$ number of $\zeta_i=1$, there exists a subset $\mathcal{S}_t^{M_t} \subseteq \mathcal{O}_t$ with $M_t$ elements. For example, in the case of $N_p=3$, the set $\mathcal{O}_t$ can be represented as $\mathcal{O}_t = \{[1, 0, 0]^\top, [0, 1, 0]^\top, [0, 0, 1]^\top\}$. If the human drawing observation is $\mathbf{o}_t^h = [1, 1, 0]^\top$, the corresponding subset would be $\mathcal{S}_t^{M_t} = \{[1, 0, 0]^\top, [0, 1, 0]^\top\}$. Here, we assume no correlation between elements within the subset $\mathcal{S}_t^{M_t}$ (i.e., the naive Bayes assumption~\cite{svensen2007pattern}). Then, $\mathcal{S}_t^{M_t}$ can be interpreted in two possibilities; the target may be located here, and it may not be located here. The underlying assumption in our human observations is as follows:
\smallskip
\begin{assumption} \label{assum::drawing}
    When a human encloses a region ($M_t\geq1$), particles are selected independently in a sequential manner.
\end{assumption}

\smallskip
Under Assumption~\ref{assum::drawing}, the likelihood function for the human drawing can be represented using the rule of the product,
\begin{align}
\label{eq::Human_likelihood_multi2}
     &p(\mathbf{o}_t^h|\mathbf{x}_t, a^h_t; \mathbf{L}) = \prod_{ \mathbf{o}_t^{M_t} } p(\mathbf{o}_t^{M_t}|\mathbf{x}_t, a^h_t; \mathbf{L}) \nonumber \\
    &\quad\qquad~~= (a^h_t)^{\delta (\mathbf{x}_t, \mathbf{o}_t^{M_t})} \times (1-a^h_t)^{\bigl(M_t-\delta (\mathbf{x}_t, \mathbf{o}_t^{M_t})\bigl)},
\end{align}
where we introduce new notations $\mathbf{o}_t^{M_t} \in \mathcal{S}_t^{M_t}$ to avoid confusion with notation $\mathbf{o}_t^h$. The likelihood function can be interpreted as representing the probability of success of human detection for an enclosed set of $M_t$ particles. Also, it has the same form as the Geometric distribution, and when coupled with the Beta distribution as the conjugate prior~\cite{svensen2007pattern}, it allows \emph{closed-form} updates of human detection reliability $a^h_t$. It will be discussed in Section~\ref{sec::BayesLearning}. 

\section{Joint Bayesian Learning for Target Localization and Human Parameters} \label{sec::BayesLearning}

This section introduces a joint Bayesian learning approach for target localization and human parameters estimation, aiming to estimate the joint posterior distribution $p(\mathbf{p}_t,a^h_t|\mathbf{o}_{1:t})$ conditioned on all measurements. The joint posterior distribution is factorized by\footnotemark{}\footnotetext{We rely on Bayes' theorem and the Chain Rule to write: \\$~~~~~\quad p(a,b|c)=\frac{p(a,b,c)}{p(c)}=\frac{p(a,b,c)}{p(b,c)} \cdot \frac{p(b,c)}{p(c)}=p(a|b,c)p(b|c).$}
\begin{align}
    p(\mathbf{p}_t,a^h_t|\mathbf{o}_{1:t})= p(a^h_t|\mathbf{p}_t, \mathbf{o}_{1:t}) p(\mathbf{p}_t|\mathbf{o}_{1:t}), \nonumber
\end{align}
where $p(\mathbf{p}_t|\mathbf{o}_{1:t})$ is the distribution of the position of the target, and $p(a^h_t|\mathbf{p}_t, \mathbf{o}_{1:t})$ is the distribution of the human parameter given the target position The main idea is to perform recursive updates on two distributions. First, we update the target position distribution based on the human parameter distribution. Next, we update the human parameter distribution by marginalizing the target position distribution. In what it follows $\propto$ indicates proportionality.

\smallskip
\emph{Bayesian Learning for Target Localization}: Given the transition model and the likelihood models, the distribution of the target position can be computed analytically in a closed form. The following result gives the distribution of the target position.
\begin{lemma} \label{lem::target}{(HMM based distribution of the target position). Let Assumption~\ref{assum::cond} hold. Consider the finite sample space $\mathbf{x}^i_t \in \mathbf{x}_t$ and $\mathbf{p}^i_t=\mathbf{L}\mathbf{x}^i_t$. Then, $q^i_{t|t}$ of \eqref{eq::Post_Target_Approx} is computed by
\begin{align}
\label{eq::Post_target}
    q^i_{t|t} = \eta \prod_{u \in \mathcal{U}} p(\mathbf{o}_t^u|\mathbf{p}_t^i)^{w_u} \prod_{h \in \mathcal{H}}  p(\mathbf{o}_t^h|\mathbf{x}_t^i; \mathbf{L})^{w_h} q^i_{t|t-1},
\end{align}
where 
\begin{align}\label{eq::marginal_human}
    &p(\mathbf{o}_t^h|\mathbf{x}_t^i; \mathbf{L})^{w_h}= \\
    &\qquad \frac{\mathsf{B}\bigl(w_h\delta (\mathbf{x}_t^i, \mathbf{o}_t^{M_t})\!+\!\alpha_t,w_h\bigl(M_t\!-\!\delta (\mathbf{x}_t^i, \mathbf{o}_t^{M_t})\bigl)+\beta_t\bigl)}{\mathsf{B}(\alpha_t,\beta_t)},\!\!\nonumber
\end{align}
with the weight parameters $w_u, w_h \in [0,1] \subseteq \real$ satisfying $\sum_{h\in\mathcal{H}} w_h+\sum_{u\in\mathcal{U}} w_u = 1$.\footnotemark{}\footnotetext{The weights $w_u$ and $w_h$ determine the relative trust assigned to each sensor. The weight assignment allows average consensus between autonomous sensors and human operators. By adjusting the weight parameters, we can balance the contribution of each sensor in multi-sensor fusion and align it with the state transition model.} 
}\end{lemma}
\begin{proof}
    Recall the two update steps~\eqref{eq::Post_update}. In the prediction~\eqref{eq::Post_update1}, $q^i_{t|t-1}$ is computed. In the correction~\eqref{eq::Post_update2}, under Assumption~\ref{assum::cond}, the likelihood $p(\mathbf{o}_t|\mathbf{p}_t^i) = p(\mathbf{o}_t^u,\mathbf{o}_t^h|\mathbf{p}_t^i)$ aggregates to $p(\mathbf{o}_t^u,\mathbf{o}_t^h|\mathbf{p}_t^i) = p(\mathbf{o}_t^u|\mathbf{p}_t^i)p(\mathbf{o}_t^h|\mathbf{p}_t^i) = \prod_{u \in \mathcal{U}} p(\mathbf{o}_t^u|\mathbf{p}_t^i)^{w_u} \prod_{h \in \mathcal{H}}  p(\mathbf{o}_t^h|\mathbf{x}_t^i; \mathbf{L})^{w_h}$. Next, we compute $p(\mathbf{o}_t^h|\mathbf{x}_t^i; \mathbf{L})^{w_h}$ from  $p(\mathbf{o}_t^h|\mathbf{x}_t^i; \mathbf{L})^{w_h}= \int_0^1 \!\!p(\mathbf{o}_t^h|\mathbf{x}_t^i, a^h_t; \mathbf{L})^{w_h} \times p(a^h_t)\text{d}a^h_t$. By substituting for $p(a^h_t)$ from~\eqref{eq::Human_param_dist_density}, this integral can be analytically computed, leading to~\eqref{eq::marginal_human}. And $\eta$ is explicitly computed to satisfy condition $\sum\nolimits_{i=1}^{N_p}\prod_{u \in \mathcal{U}} p(\mathbf{o}_t^u|\mathbf{p}_t^i)^{w_u} \times \prod_{h \in \mathcal{H}}  p(\mathbf{o}_t^h|\mathbf{x}_t^i; \mathbf{L})^{w_h} q^i_{t|t-1}=1$. Thus, for each $\mathbf{p}_t^i$, $q^i_{t|t}$ is computed, leading to $p(\mathbf{p}_t|\mathbf{o}_{1:t})$ in the form of~\eqref{eq::Post_Target_Approx}.
\end{proof}

\begin{algorithm}[t]
\caption{Joint Bayesian Online Learning}
\label{alg:JBOL}
\begin{algorithmic}[1]
    \STATE Init: $\{q^i_0\}_{i=1}^{N_p} \sim p(\mathbf{p}_0)$, $a^h_0\sim\mathsf{Beta}(\alpha_0^h,\beta_0^h)$, $w_h, w_u$. 
    \FOR{$t=1,2,\dots$}
        \STATE $\{q^i_{t|t-1}\}_{i=1}^{N_p}$ $\stackrel{\eqref{eq::Post_update1}}{\leftarrow}\{q^i_{t-1|t-1}\}_{i=1}^{N_p}$ or $\{q^i_{t-1}\}_{i=1}^{N_p}$ if $t=1$.
        \IF{Human observation is applied with $M_t\geq1$}
            \STATE $\{q^i_{t|t}\}_{i=1}^{N_p}$   $\stackrel{\eqref{eq::Post_target}}{\leftarrow}(\alpha_t^h,\beta_t^h, \mathbf{o}_t^u,\mathbf{o}_t^h,w_h,w_u,\{q^i_{t|t-1}\}_{i=1}^{N_p})$.
            \STATE $(\alpha_t^h,\beta_t^h)\!\stackrel{\text{MM}}{\leftarrow}\!$   $p(a^h_t|\mathbf{o}_{1:t})\stackrel{\eqref{eq::Post_Human}}{\leftarrow}(\{q^i_{t|t}\}_{i=1}^{N_p},\mathbf{o}_t^h,w_h)$.
        \ELSE 
            \STATE $\{q^i_{t|t}\}_{i=1}^{N_p} \stackrel{\eqref{eq::Post_target}}{\leftarrow}(\mathbf{o}_t^u,w_u,\{q^i_{t|t-1}\}_{i=1}^{N_p})$. 
        \ENDIF

    \ENDFOR
    \STATE Out: MMSE result $\hat{\mathbf{p}}_{1:t}$, human parameters $(\alpha_{1:t}^h,\beta_{1:t}^h)$. 
\end{algorithmic}
\end{algorithm}

\medskip
\emph{Bayesian Learning for Human Parameters}:
Given the posterior distribution of the target position $p(\mathbf{p}_t|\mathbf{o}_{1:t})$, $p(a^h_t|\mathbf{o}_{1:t})$ is computed via marginalization of $\mathbf{p}_t$, given by $\int_{\Theta} p(a^h_t|\mathbf{p}_t, \mathbf{o}_{1:t}) p(\mathbf{p}_t|\mathbf{o}_{1:t}) \text{d}\mathbf{p}_t$. The following result gives the distribution of the human parameter (detection reliability). 

\begin{lemma}{(Distribution of the human detection reliability). Given the $N_p$ number of~\eqref{eq::Post_target}, $p(a^h_t|\mathbf{o}_{1:t})$ is computed as
\begin{align}
    \label{eq::Post_Human}
    p(a^h_t|\mathbf{o}_{1:t})=\,&  q_t^s \cdot \mathsf{Beta}(w_h+\alpha_t, w_h(M_t - 1) + \beta_t) \nonumber \\
    &+\Bigl(1 - q_t^s\Bigl) \cdot \mathsf{Beta}(\alpha_t, w_h M_t +\beta_t),
\end{align}
where the $q_t^s$ denotes the sum of the weight $q^i_{t|t}$ when $\mathbf{x}^i_t = \mathbf{o}_t^{M_t}$ (i.e., $q_t^s = \sum_{\{\mathbf{x}^i_t | \mathbf{x}^i_t =  \mathbf{o}_t^{M_t}\}}q^i_{t|t}$).
}
\end{lemma}
\begin{proof}
The distribution of the human parameter can be represented via the marginalization of the target position as
\begin{align}
\label{eq::Appendix_human_para}
    &p(a^h_t|\mathbf{o}_{1:t}) = \sum\nolimits_{i=1}^{N_p} p(a^h_t|\mathbf{p}_t^i, \mathbf{o}_{1:t}) \cdot q^i_{t|t}
    \nonumber \\
    &= \sum\nolimits_{i=1}^{N_p} p(\mathbf{o}_t^h|\mathbf{x}_t^i, a^h_t;\mathbf{L})^{w_h} p(a^h_t|\alpha_t,\beta_t) \cdot q^i_{t|t}, \nonumber
\end{align}
where $p(\mathbf{o}_t^h|\mathbf{x}^i_t, a^h_t;\mathbf{L})^{w_h}$ and $p(a^h_t|\alpha_t,\beta_t)$ show a conjugate prior relationship. Therefore, for given each $\mathbf{p}^i_t (=\mathbf{L}\mathbf{x}^i_t)$, a new Beta distribution can be obtained by $p(\mathbf{o}_t^h|\mathbf{x}^i_t, a^h_t;\mathbf{L})^{w_h} \cdot p(a^h_t|\alpha_t,\beta_t) = \mathsf{Beta}(w_h\delta (\mathbf{x}^i_t, \mathbf{o}_t^{M_t})+\alpha_t, w_h\bigl(M_t-\delta (\mathbf{x}^i_t, \mathbf{o}_t^{M_t})\bigl)+\beta_t)$. Then, it can be represented as a weighted sum of two Beta distributions as given by
\begin{align}
    &=\!\! \sum_{i=1}^{N_p} \! q^i_{t|t}\! \cdot \mathsf{Beta}\bigl(w_h\delta (\mathbf{x}^i_t, \mathbf{o}_t^{M_t})\!+\!\alpha_t, \!w_h\bigl(M_t\!\!-\!\delta (\mathbf{x}^i_t, \mathbf{o}_t^{M_t})\bigl)+\beta_t\bigl) \nonumber \\
    &=\!\!\!\!\!\!  \sum_{\{\mathbf{x}^i_t | \mathbf{x}^i_t =  \mathbf{o}_t^{M_t}\}}q^i_{t|t} \cdot  \mathsf{Beta}(w_h+\alpha_t, w_h(M_t - 1) + \beta_t) \nonumber \\
    &\quad~~~~~~~+\biggl(1 - \!\!\sum_{\{\mathbf{x}^i_t | \mathbf{x}^i_t =  \mathbf{o}_t^{M_t}\}}q^i_{t|t}\biggl) \cdot \mathsf{Beta}(\alpha_t, w_h M_t +\beta_t) \nonumber \\
    &= q_t^s \cdot \mathsf{Beta}(w_h+\alpha_t, w_h(M_t - 1) + \beta_t) \nonumber \\
    &\qquad\quad~+\Bigl(1 - q_t^s\Bigl) \cdot \mathsf{Beta}(\alpha_t, w_h M_t +\beta_t). \nonumber
\end{align}
Thus, $p(a^h_t|\mathbf{o}_{1:t})$ become~\eqref{eq::Post_Human}. 
\end{proof}

\begin{figure}[t]
    \centering
    \includegraphics[width=0.28\textwidth]{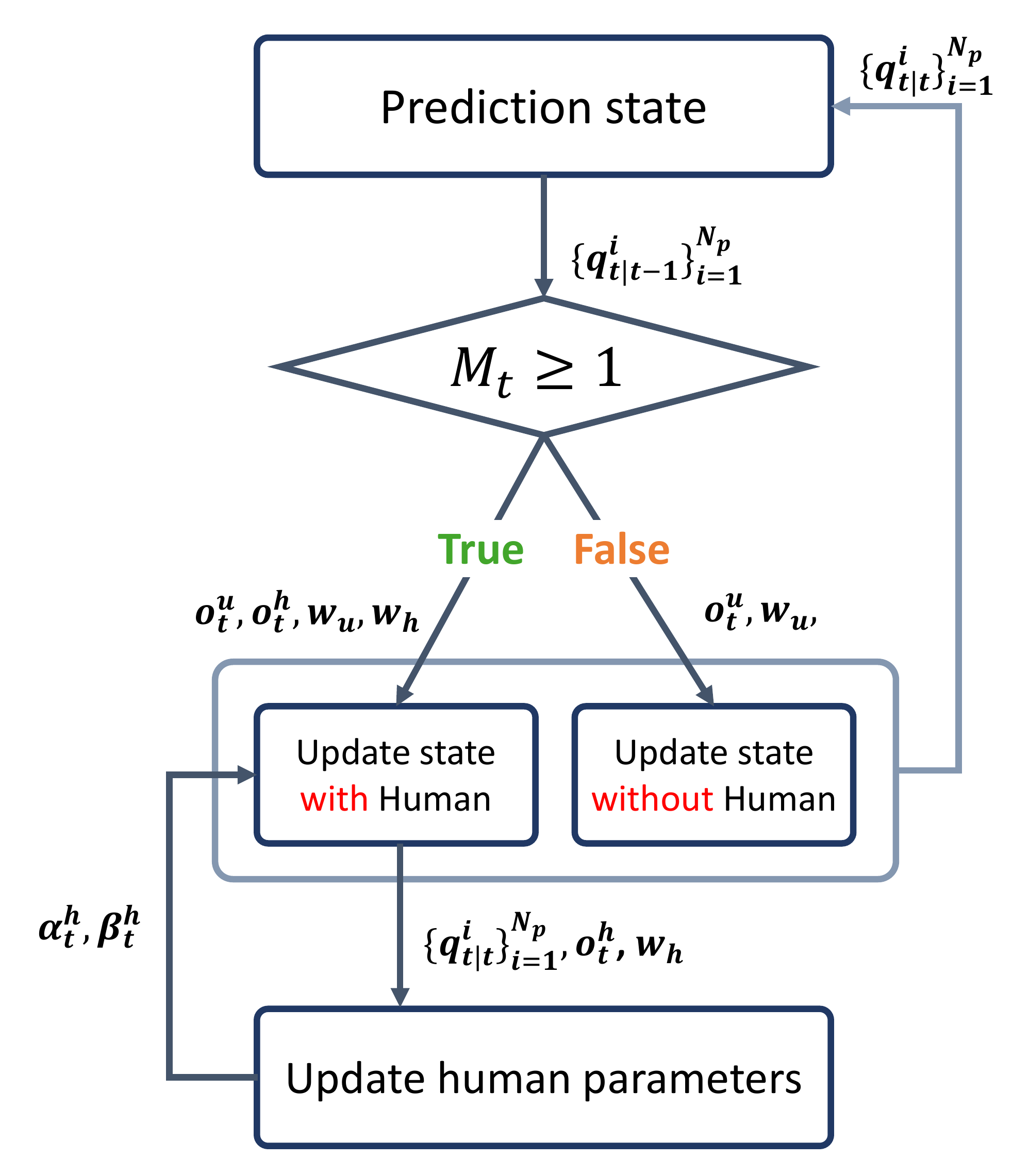}
    \caption{{\small A high level flow diagram of the proposed framework.}}
    \label{fig:FD}
\end{figure}

\smallskip
The updated distribution of the human parameter is represented as a weighted sum of two Beta distributions, which, in general, differs from a single Beta distribution~\cite{wackerly2014mathematical}. However, when the two Beta distributions share identical parameters, the resulting distribution is simplified into a single Beta distribution.

To improve computation efficiency, we propose using a moment-matching method (MM) instead of computationally demanding approaches~\cite{svensen2007pattern}. To preserve the conjugate prior form, we approximate the weighted sum of two Beta distributions as a single Beta distribution.\footnotemark{}\footnotetext{In practice, since there is a minimal significant difference between the two Beta distributions $(0 \leq w_h \leq 1)$, it is highly possible that the weighted sum of two Beta distributions will closely resemble a single Beta distribution.} We approximate the resulting distribution of $p(a^h_t|\mathbf{o}_{1:t})$ as a Beta distribution $a^h_t \sim \mathsf{Beta}(\alpha_t^{\star}, \beta_t^{\star})$ by employing the first two moments, the mean $\mathbb{E}[a^h_t]$ and variance $\text{var}[a^h_t]$. As a result, we obtain two equations given by: $\frac{\alpha_t^{\star}}{\alpha_t^{\star}+\beta_t^{\star}} = \frac{q_t^s \cdot(w_h+\alpha_t) + (1-q_t^s)\cdot\alpha_t}{w_h M_t+\alpha_t+\beta_t}$ and $\frac{\alpha_t^{\star}\beta_t^{\star}}{(\alpha_t^{\star}+\beta_t^{\star})^2\cdot(\alpha_t^{\star}+\beta_t^{\star}+1)} = \frac{q_t^s \cdot~(w_h+\alpha_t) \cdot (w_h(M_t - 1) + \beta_t) + (1-q_t^s)\cdot~\alpha_t \cdot (w_h M_t +\beta_t)}{(w_h M_t+\alpha_t+\beta_t)^2\cdot(w_h M_t+\alpha_t+\beta_t+1)}$. From these two equations, we can show that a unique $(\alpha_t^{\star}, \beta_t^{\star})$ can be computed numerically. This approach provides significant advantages for recursively updating human parameters with~\eqref{eq::Human_likelihood_multi2}, allowing real-time target localization. Algorithm~\ref{alg:JBOL} and a flow diagram in Fig.~\ref{fig:FD} summarize the proposed method. Additionally, we obtain the following result on the asymptotic human reliability.
\begin{theorem} (Asymptotic human reliability). When human observation is applied with $M_t\geq1$ at each time $t$, $a^h_t$ converges in probability, i.e., there exists $a^h$ such that, for any $\epsilon>0$, $\lim_{t\rightarrow\infty}\text{Pr}(|a^h_t-a^h|<\epsilon)=1$.
\end{theorem}
\begin{proof}
    When human observation is applied with $M_t\geq1$ at each time $t$, by the law of large numbers, the expected reliability $\lim_{t\rightarrow\infty}\mathbb{E}[a^h_t]=\lim_{t\rightarrow\infty}\frac{\alpha_t}{\alpha_t+\beta_t} = \lim_{t\rightarrow\infty}\frac{\sum_{k=1}^t w_h q_k^s + \alpha_0}{\sum_{k=1}^t w_h M_k+\alpha_0+\beta_0}$ where $\alpha_0$ and $\beta_0$ are initial values. Given $0\leq w_h\leq1$, $0\leq q_t^s \leq1$, and $1\leq M_t\leq N_p$, there exists an upper bound $\sum_{k=1}^t w_h q_k^s + \alpha_0 \leq w_h t + \alpha_0$. Similarly, $\sum_{k=1}^t w_h M_k+\alpha_0+\beta_0 \leq w_h N_p t +\alpha_0+\beta_0$. Then $\lim_{t\rightarrow\infty}\mathbb{E}[a^h_t]$ is bounded. Also, for any $\epsilon>0$, by the Markov inequality, $\lim_{t\rightarrow\infty}\text{Pr}(|a^h_t-a^h|>\epsilon) \leq \lim_{t\rightarrow\infty}\frac{1}{\epsilon^2}\mathbb{E}[(a^h_t-a^h)^2]=0$ where $\lim_{t\rightarrow\infty}\text{var}[a^h_t]=0$ as $\lim_{t\rightarrow\infty}(\alpha_t+\beta_t)=\infty$, derived from $\lim_{t\rightarrow\infty}\sum_{k=1}^t M_k=\infty$. Hence, $\lim_{t\rightarrow\infty}\text{Pr}(|a^h_t-a^h|<\epsilon)=1$.
\end{proof}

\begin{figure}[t]
    \centering
    \includegraphics[width=0.47\textwidth]{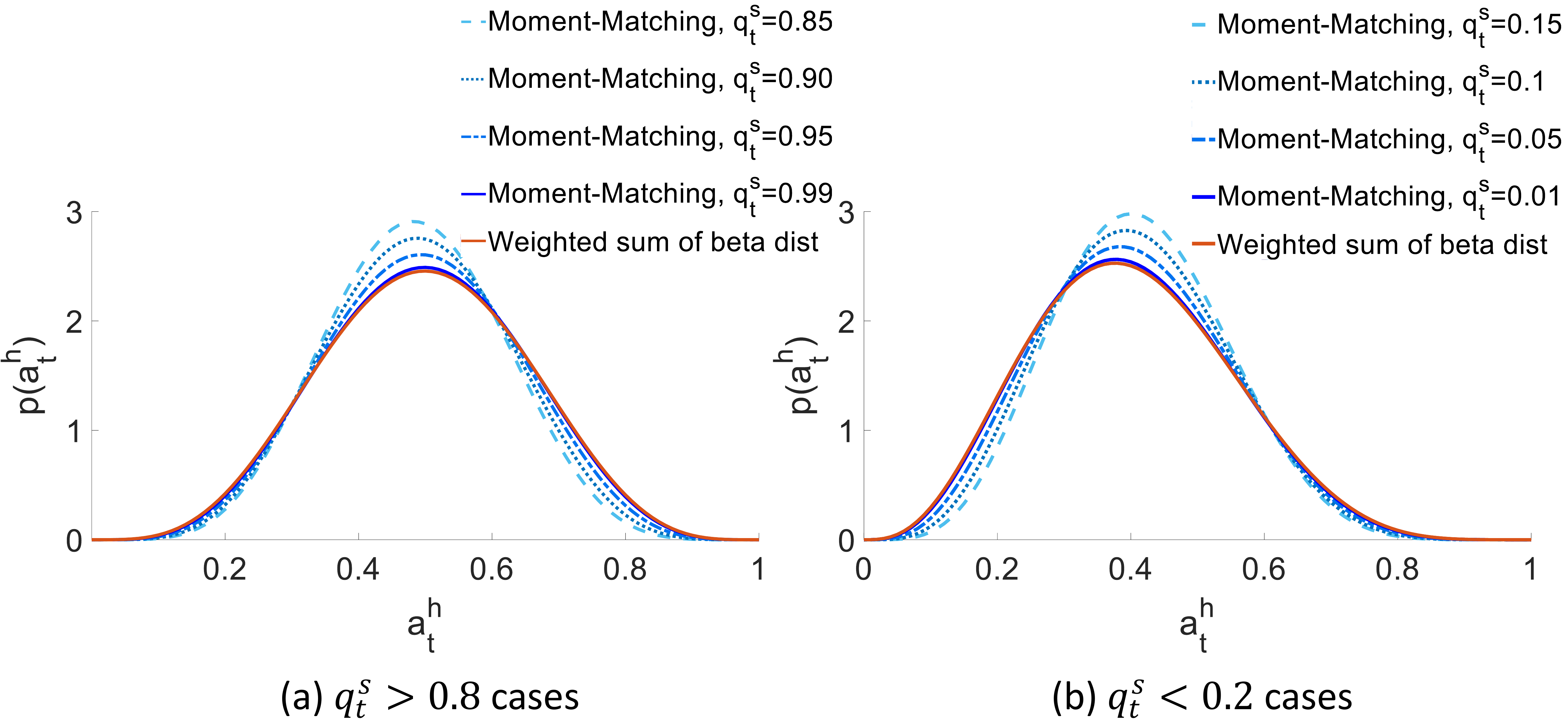}
    \caption{{\small Illustration of how the moment-matching result approaches the weighted sum of two Beta distributions.}}
    \label{fig:Moment_Matching}
\end{figure}

\medskip
\begin{remark}
    In practice, the accuracy of the result of the moment-matching method is significantly based on $q_t^s$. When human operators attentively make accurate drawing observations, the value of $q_t^s$ increases significantly, i.e., $q_t^s > 0.8$. In contrast, substantial errors by human operators lead to a significant decrease in the value of $q_t^s$, i.e., $q_t^s < 0.2$. As a result, the moment-matching outcomes exhibit a high level of accuracy; see Fig.~\ref{fig:Moment_Matching}. 
\end{remark}

\smallskip
Finally, human-assisted autonomy localization can yield MMSE estimation variance equal to or lower than that of only autonomous sensor-based localization.

\medskip
\begin{theorem} (Human-assisted localization). Let Assumption~\ref{assum::cond} holds. The fusion of human observations ($M_t\geq1$) in the updated posterior can result in a variance that is either equal to or lower than that of only autonomous sensor-based localization, i.e., $\mathbb{E}[(\mathbf{p}^i_t-\hat{\mathbf{p}}_t)^{\top}(\mathbf{p}^i_t-\hat{\mathbf{p}}_t)|\mathbf{o}_t^u,\mathbf{o}_t^h] \leq \mathbb{E}[(\mathbf{p}^i_t-\hat{\mathbf{p}}_t)^{\top}(\mathbf{p}^i_t-\hat{\mathbf{p}}_t)|\mathbf{o}_t^u]$.
\end{theorem}
\begin{proof}
    To evaluate variance, we compute the mean $\hat{\mathbf{p}}_t^1:=\mathbb{E}[\mathbf{p}_t|\mathbf{o}_t^u,\mathbf{o}_t^h]$, which incorporates human observation, and $\hat{\mathbf{p}}_t^2:=\mathbb{E}[\mathbf{p}_t|\mathbf{o}_t^u]$ which is without human observation. To simplify the notation, we assume $|\mathcal{H}|=|\mathcal{U}|=1$ where $|\cdot|$ indicates cardinality of a set. Under Assumption~\ref{assum::cond}, we express the means as $\hat{\mathbf{p}}_t^1=\eta_1\sum\nolimits_{i=1}^{N_p}(l_t^h)^i (l_t^u)^i q^i_{t|t-1} \mathbf{p}_t^i$ and $\hat{\mathbf{p}}_t^2=\eta_2\sum\nolimits_{i=1}^{N_p} (l_t^u)^i q^i_{t|t-1} \mathbf{p}_t^i$, where $(l_t^h)^i=p(\mathbf{o}_t^h|\mathbf{x}_t^i; \mathbf{L})^{w_h}$ and $(l_t^u)^i=p(\mathbf{o}_t^u|\mathbf{p}_t^i)^{w_u}$. Here, $\eta_1$ ensures $\sum\nolimits_{i=1}^{N_p}(l_t^h)^i (l_t^u)^iq^i_{t|t-1}=1$, and $\eta_2$ ensures $\sum\nolimits_{i=1}^{N_p}\\ (l_t^u)^i q^i_{t|t-1}=1$. Next, we evaluate the variances $\mathbb{E}[(\mathbf{p}^i_t-\hat{\mathbf{p}}_t^1)^{\top}(\mathbf{p}^i_t-\hat{\mathbf{p}}_t^1)|\mathbf{o}_t^u,\mathbf{o}_t^h] = \eta_1\sum\nolimits_{i=1}^{N_p} (l_t^h)^i (l_t^u)^i q^i_{t|t-1}(\mathbf{p}^i_t-\hat{\mathbf{p}}_t^1)^{\top}(\mathbf{p}^i_t-\hat{\mathbf{p}}_t^1)$ and $\mathbb{E}[(\mathbf{p}^i_t-\hat{\mathbf{p}}_t^2)^{\top}(\mathbf{p}^i_t-\hat{\mathbf{p}}_t^2)|\mathbf{o}_t^u] = \eta_2\sum\nolimits_{i=1}^{N_p} (l_t^u)^i q^i_{t|t-1}(\mathbf{p}^i_t-\hat{\mathbf{p}}_t^2)^{\top}(\mathbf{p}^i_t-\hat{\mathbf{p}}_t^2)$. If $\hat{\mathbf{p}}_t=\hat{\mathbf{p}}_t^1 = \hat{\mathbf{p}}_t^2$ and human observations are applied, it reduces the contribution of states $\mathbf{p}^i_t$ with high variance. In other words, $(l_t^h)^i$ from \eqref{eq::marginal_human} in Lemma~\ref{lem::target} is chosen as: $\eta_1\sum\nolimits_{i=1}^{N_p}(l_t^h)^i (l_t^u)^i q^i_{t|t-1} (\mathbf{p}^i_t-\hat{\mathbf{p}}_t)^{\top}(\mathbf{p}^i_t-\hat{\mathbf{p}}_t) \leq \eta_2\sum\nolimits_{i=1}^{N_p}(l_t^u)^i q^i_{t|t-1} (\mathbf{p}^i_t-\hat{\mathbf{p}}_t)^{\top}(\mathbf{p}^i_t-\hat{\mathbf{p}}_t)$. Thus, this implies that $\mathbb{E}[(\mathbf{p}^i_t-\hat{\mathbf{p}}_t)^{\top}(\mathbf{p}^i_t-\hat{\mathbf{p}}_t)|\mathbf{o}_t^u,\mathbf{o}_t^h] \leq \mathbb{E}[(\mathbf{p}^i_t-\hat{\mathbf{p}}_t)^{\top}(\mathbf{p}^i_t-\hat{\mathbf{p}}_t)|\mathbf{o}_t^u]$.
\end{proof}

\smallskip
\section{Simulation Study}
\label{sec::simulation}
\begin{figure}[!t]
    \centering
    \includegraphics[width=0.47\textwidth]{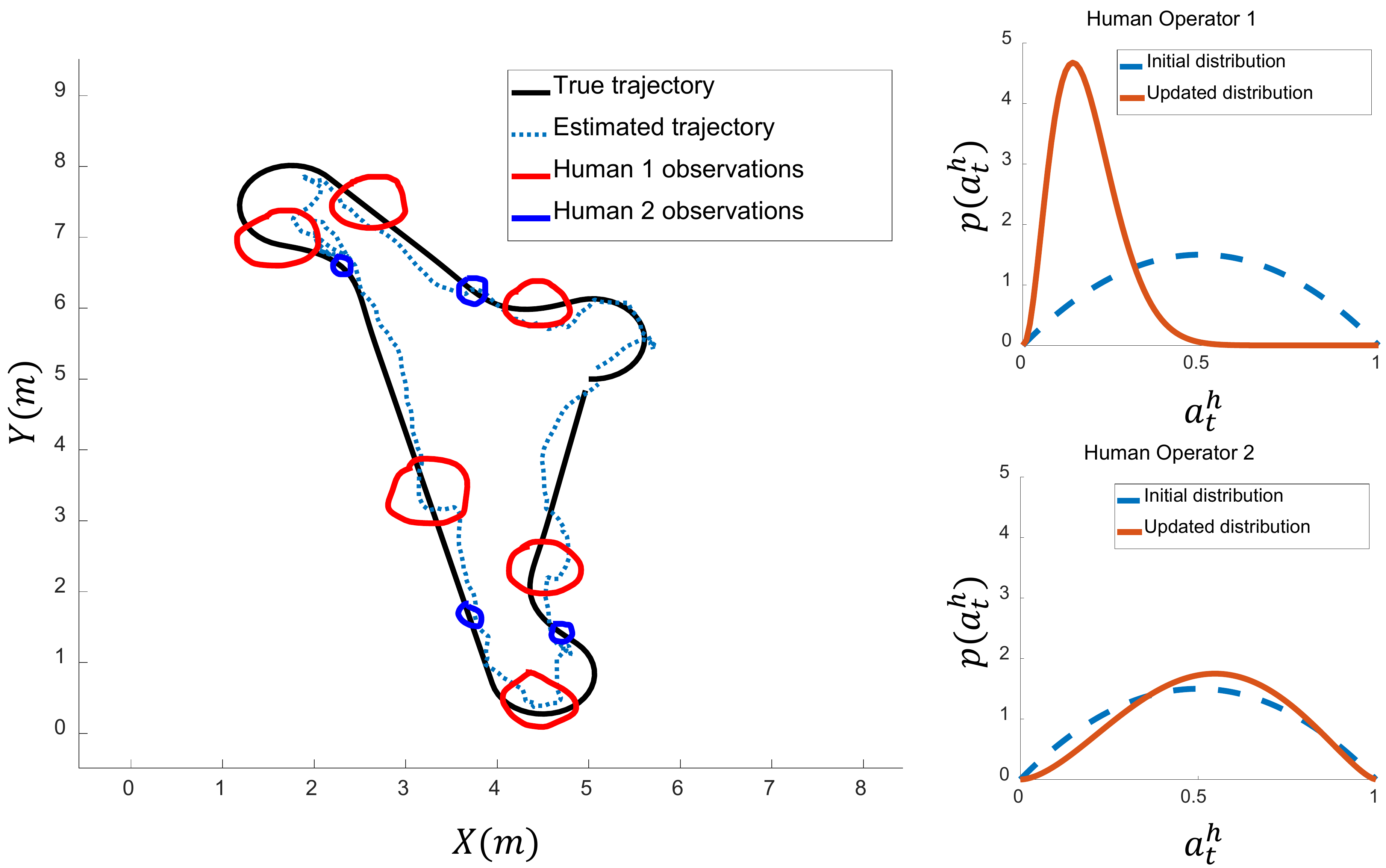}
    \caption{{\small The proof-of-concept simulation. (Left) The red drawings represent contributions from human operator $1$, while the blue drawings reflect input from human operator $2$. (Right) Both the initial distribution and the updated distribution of the human operator's detection reliability.}}
    \label{fig:Sim_result}
\end{figure}

We assess the effectiveness of our human-assisted autonomy sensor fusion-based localization algorithm through a simulated scenario. The scenario involves the collaborative efforts of three UAVs (autonomous sensors) and two human operators to accurately localize a mobile target. During simulations, the target is dynamically rendered on the screen, providing a visual representation of its movement. To localize the target, we use a constant velocity motion model that is widely used in tracking problems~\cite{bewley2016simple} as $$ p(\mathbf{p}_t|\mathbf{p}_{t-1}) = \mathcal{N}(\mathbf{p}_t;\mathbf{p}_{t-1} + \mathbf{v}_{t-1} T_s, T_s^2\sigma_p^2/2\mathbf{I}),$$ where $\mathbf{v}_{t} \sim \mathcal{N}(\mathbf{v}_{t};\mathbf{v}_{t-1},T_s\sigma_p^2 \mathbf{I})$ and $\mathbf{I}$ is the identity matrix according to the dimension. Here, we set $\sigma_p = 0.5~(m)$, $T_s = 0.1(s)$, and $\mathbf{v}_0 = [1~1]^{\top}(m/s)$. For the autonomous sensor, we apply the likelihood model from stereo vision with the detection algorithm~\cite{farhadi2018yolov3,hartley2003multiple} as $p(\mathbf{o}_t^u|\mathbf{p}_t) = \mathcal{N}(\mathbf{o}_t^u; r_t, \sigma_u^2) \Gamma(D_t|p_d)$ where $r_t = ||\mathbf{c}_{u,t} - \mathbf{p}_t^+||_2 \in \real_{\geq 0}$ is the range measurement and $\Gamma(D_t|p_d)$ is the detection event model, e.g., $\Gamma(D_t=1|p_d)=1$ (detected) or $\Gamma(D_t=0|p_d)=0$ (undetected), with success probability $p_d$. Here, we set $\sigma_u = 0.05~(m)$ and two different values $p_d$. The human operators participate by using a stylus pen to encircle the target's estimated location on the display screen. In our simulation setting, the two human operators initialize their initial detection reliability, characterized by the same parameters $a_1^{h_1}, a_1^{h_2} \sim \mathsf{B}(2, 2)$ (i.e., mediocre level), as indicated by the blue dotted line in Fig.~\ref{fig:Sim_result} (right). The initial position of the target is known prior. In the discrete sample space, we distribute the $N_p = 400$ particles over the region of interest $\Theta = [0,10]~(m) \times [0,10]~(m)$. We assigned the weights $w_h = 2/13, w_u = 3/13$ based on the number of autonomous sensors and human operators.

\begin{table}[t]
{\footnotesize
\caption{Comparison of RMSE in $100$ MC Simulations}
\label{Table_1}\setlength{\tabcolsep}{3pt}\renewcommand{\arraystretch}{0.9}
\begin{center}
\begin{tabular}{|c | c | c|}
\hline
& Human + Autonomous & Autonomous \\
& RMSE [m] & RMSE [m] \\
\hline
$p_d (0.8)$ & $\mathbf{0.375}$  & $0.390$  \\
\hline
$p_d(0.8)$ + $10$ faults & $\mathbf{0.380}$ & $0.401$ \\
\hline
$p_d(0.6)$ & $\mathbf{0.410}$ & $0.427$  \\ 
\hline
$p_d(0.6)$ + $10$ faults & $\mathbf{0.412}$ & $0.433$  \\
\hline
3 detection failures & $\mathbf{0.438}$  & $0.517$  \\
\hline
5 detection failures & $\mathbf{0.454}$ & $0.586$ \\
\hline
7 detection failures & $\mathbf{0.480}$ & $0.647$  \\ 
\hline
\end{tabular}
\end{center}
}
\end{table}

We evaluated the localization accuracy in two cases across four different scenarios, as shown in Table~\ref{Table_1}, using the root mean square error (RMSE) metric: (1) autonomous sensor-based localization and (2) autonomous + human sensor fusion. In the table, `10 faults' refers to a scenario in which the autonomous sensors each had $10$ occasions that their measurements were faulty due to the existence of $+1~m$ bias in their range measurements. In the first four scenarios reported in Table~\ref{Table_1}, the hard sensor measurements take place at each time step with a probability specified by $p_d$. In the next three scenarios reported in Table~\ref{Table_1}, `$\times$ detection failures' refers to a scenario in which the hard sensors failed to detect the target in $\times$ number of instances for a duration of $2$ to $4$ seconds. To compare the update process of human detection reliability distributions, two human operators provided observations on different-sized drawings (large vs small), as shown in Fig.~\ref{fig:Sim_result} (left). In Table~\ref{Table_1}, although there are a few observations ($10$ observations) compared to autonomous sensors, the integration of human drawing data introduces a compensatory mechanism that effectively mitigates estimation errors in sensors, including those with detection failures and faulty measurements. The impact of human assistance is more pronounced when the autonomous sensors are disconnected for an extended period of time as shown in the last three scenarios in Table~\ref{Table_1}.

In Fig.~\ref{fig:Sim_result} (right), the updated distributions of detection reliability are depicted for two human operators. Human operator $1$ experiences reduced reliability due to the larger size of the drawings, even when the observations are accurate. In contrast, human operator $2$ benefits from increased reliability due to the smaller size of the correct drawings. These results effectively reflect the detection reliability based on observations and estimation results. Regarding the moment-matching method, we examine the value of $q_t^s$ throughout the simulations. In this simulation, given the reasonable observations provided by human operators, we can observe that $q_t^s$ reaches a threshold of at least $0.9995$. Consequently, this allows us to derive an accurate updated distribution of human reliability. 

\section{Conclusions}
\label{sec::Con}
This paper considered a human-assisted autonomy sensor fusion for dynamic target localization in a Bayesian framework. 
We devised a novel human drawing observation model to provide spatial information, addressing the limitations of autonomous sensors. 
The design of the human drawing observation took into account both human detection reliability and inherent uncertainties. With this model, we proposed a joint Bayesian learning approach that encompasses both target localization and dynamic updating of human detection reliability in a computationally efficient manner. Lastly, we demonstrated improved target localization outcomes, paralleled by the updated distribution of human detection reliability. In future work, our focus is on addressing the assumption of non-correlation between sample spaces in human drawing observations. We also plan to make modeling efforts to capture the spatial information extracted from the intricate shapes of human drawings. Taking these factors into account, the precise quantification of human observation will improve the accuracy of target localization.

\bibliographystyle{ieeetr}
\bibliography{bib/alias.bib, bib/Reference.bib}
\end{document}